\documentclass[10pt,twocolumn,letterpaper]{article}
\usepackage[T1]{fontenc}    

\usepackage{wacv2013-author-kit/latex/wacv}
\usepackage{times}
\usepackage{epsfig}
\usepackage{graphicx}
\usepackage{amsmath}
\usepackage{amssymb}
\usepackage{tabularx}

\usepackage{algorithm}
\usepackage{algorithmic}
\usepackage{multirow}

\usepackage{graphicx}
\usepackage{epstopdf}

\usepackage{caption}
\usepackage{subcaption}
\usepackage{fancyhdr}

\fancypagestyle{firstpage}{%
  \fancyhf{}

  \fancyfoot[C]{\thepage}
  \fancyfoot[L]{\footnotesize This paper was originally published in\\2013 IEEE Workshop on Applications of Computer Vision (WACV).\\
  \space DOI: 10.1109/WACV.2013.6475023}
}

\input{perso.sty}

\newlength{\subfigheight}
\usepackage[pagebackref=true,breaklinks=true,letterpaper=true,colorlinks,bookmarks=false]{hyperref}

\wacvfinalcopy 

\def\wacvPaperID{74} 

\ifwacvfinal\fi
\begin{document}


\title{HotSpotter - Patterned Species Instance Recognition}

\author{Jonathan P. Crall\\
Rensselaer Polytechnic Institute\\
{\tt\small cralljp@cs.rpi.edu}
\and
Charles V. Stewart\\
Rensselaer Polytechnic Institute\\
{\tt\small stewart@cs.rpi.edu}
\and
Tanya Y. Berger-Wolf\\
University of Illinois-Chicago\\
{\tt\small tanyabw@uic.edu}\\
\and
Daniel I. Rubenstein\\
Princeton University\\
{\tt\small dir@princeton.edu}\\
\and
Siva R. Sundaresan\\
Denver Zoological Foundation\\
{\tt\small ssundare@princeton.edu}\\
}

\makeatletter
\def\@maketitle
   {
   \newpage
   \null
   \vskip .375in
   \begin{center}
      {\Large \bf \@title \par}
      \vspace*{24pt}
      {
      \large
      \lineskip .5em
      \begin{tabular}[t]{c}
         \ifwacvfinal\@author\else Anonymous WACV submission\\
         \vspace*{1pt}\\
Paper ID \wacvPaperID \fi
      \end{tabular}
      \par
      }
      \vskip .5em
      \vspace*{2pt}
   \end{center}
   }

\makeatother

\maketitle
\thispagestyle{firstpage}

\begin{abstract}
We present HotSpotter, a fast, accurate algorithm
for identifying individual animals against a labeled database.
It is not species specific and has been applied to
Grevy's and plains zebras, giraffes, leopards, and lionfish.
We describe two approaches,
both based on extracting and matching keypoints or ``hotspots''.
The first tests each new query image sequentially against each
database image, generating a score for each database image
in isolation, and ranking the results. The second, building
on recent techniques for instance recognition, matches
the query image against the database using a fast nearest
neighbor search.  It uses a competitive scoring
mechanism derived from the Local Naive Bayes Nearest
Neighbor algorithm recently proposed for category recognition.
We demonstrate results on databases of more than
1000 images, producing more accurate matches than published
methods and matching each query image in just a few seconds.
\end{abstract}

\section {Introduction}

\begin{figure}
\centering
     \includegraphics[width=\linewidth] {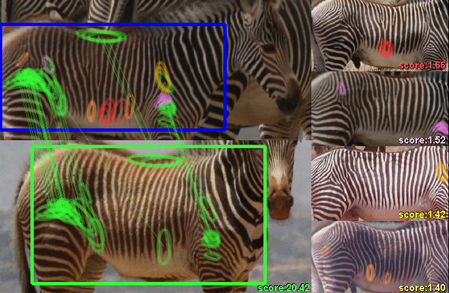}
\caption{\footnotesize{An example of HotSpotter. The ROI is placed on a
  query animal --- in this case a foal --- and a ranked list of
  matching animal images from the database is returned.  Here, the correct
  match, an image of the same animal as an older juvenile, is the best match
  (top right).  The green ellipses show the matching regions, or ``hot
  spots'', between the two images.  Images along the bottom of the
  figure show significantly lower matching scores for different animals.}}
\label{fig:onevsone}
\end{figure}

\textbf{Motivation:}
Conducting research on animal populations
requires reliable information on the position and movement of individual
animals.  Traditionally, this information has been obtained by
attaching tags and transmitters to captured animals.  These methods
do not scale well to large populations, are expensive, physically invasive, and require proximity to unwilling
subjects \cite{wsb94ElbinMicrochipIden, acmss04ZhangZebraNet}.

The widespread availability of inexpensive, good-quality digital
cameras offers the promise of an alternative approach.  Images of
animals may be taken by anyone who has a camera --- scientists and
their assistants, ecotourists, and even ordinary citizens ---
producing the potential for enormous flood of image data.
Moreover, recent cameras include both a clock and a GPS unit, allowing
each image to also record location and time.

Fundamentally, exploiting the wealth of photographic data for animal
population analysis depends on locating and recognizing the animals in
each image.  The recognition step requires comparing each animal image
to a database of pictures of animals that have already been
identified, and then either adding to the record of a
previously known animal or creating an additional record for a new
individual.  Even for small populations of 100 animals,
doing this comparison manually is tedious and error prone.  It does
not scale at all to large populations.
Clearly, computer-based methods for automatic animal identification 
are needed.

This need has spawned research on recognition of
animals ranging from penguins to zebras to whale sharks \cite{ipc10SherleyCVPenguin, jae05HolmbergAstroWhaleShark}.  Some methods
are species specific \cite{si04BradfieldFrogPhoto, aje09ShorrocksNeckNet, icmr11LahiriStripeSpotter},
while others have strived for generality \cite{fiz07SpeedSpotMatch, 11BoldgerWILDID}.
In particular, the Wild-ID algorithm of Bolger \etal~\cite{11BoldgerWILDID} employs
keypoint matching techniques
from computer vision literature \cite{ijcv04LoweSIFT} to
determine if two images show the same
animal. In our paper, we build on more recent methods from the computer 
vision literature to take an important step beyond this, creating both
an improved method for matching two images
and a fast method for matching against the entire database
without sequential image-to-image matching. 

\textbf{Problem Statement:} Our computational task is as follows.  We
are given a database of labeled images, where each label identifies
(names) the individual animal in the image. The same animal may
appear in more than one database image. We keep multiple images to
accommodate viewpoint differences and appearance changes.  Given
a novel query image, $I_Q$, and a
manually-specified rectangular region of interest (ROI) locating the
animal in the image, our goal is to
assign a label to the animal in the ROI or to decide that the animal has
not been seen before.  More practically, the goal is modified
slightly: to provide a set of potential animal labels from the
database ranked by a similarity score.
A high score should indicate a highly probable match,
while low scores should indicate improbable matches.
Figure~\ref{fig:onevsone} shows an example.

Three comments about this problem are appropriate.  First, user
interaction is required only to select the animal ROI in the query image
and to verify the labeling results.  Ongoing work will replace both of
these with more automated methods, but some user interaction will
always be necessary, both for the hardest cases and to build user
confidence in the system.
Second, for the current work, we assume the collection protocol has produced images all of one flank of the animals, avoiding the ambiguity associated with seeing two different sides of the same animal.
Third, for the experiments in this paper,
the database of animal images and labels is
static. In practice, however, the database will be dynamic.  We bias
our selection of techniques in light of this, and the system
described here is already in use with a changing image set.
However, fully addressing the issues of a efficiently searchable, large-scale, and dynamic
database of animal images is beyond the scope of this paper.

\textbf{Algorithm Overview:} We present two algorithms to solve the
animal identification problem.  A high-level summary of both is:
(a) for every image, the algorithms locate keypoints and extract associated descriptors (128-dimensional
vectors), and (b) they then determine image matches based on the comparison of these descriptors.

  The first algorithm is similar to Wild-ID
\cite{11BoldgerWILDID} in that it matches the query image
against each database image separately, sorting the database images by the
resulting similarity score in order to generate the final ranked
results.  This is our \emph{one-vs-one} matching algorithm, and
it contains several minor improvements over Wild-ID that result in a more robust
algorithm.

Our second algorithm, which we refer to as the \emph{one-vs-many}
algorithm, matches each descriptor from the query image against
\textbf{all} descriptors from the database image using a fast,
approximate nearest neighbor search data structure.  It generates scores
for each database image based on these matches, and then aggregates
the scores to produce the final, overall similarity score for each
individual (each label).  In addition to introducing fast matching to the animal
identification problem, our most important contribution here is a new
mechanism for scoring descriptor matches based on the Local Naive Bayes Nearest 
Neighbor methods \cite{cvpr11McCannLNBNN}.

\textbf{Dataset:} We test our algorithms, which are currently in use in the
field in Kenya, on five different species of animals: giraffes,
jaguars, lionfish, plains zebras and Grevy's zebras.  Our most
extensive databases are for the zebras, so these are the focus of the
experiments in this paper.

\newcommand{\plusequals}{\hspace*{1mm}+\hspace*{-1mm}=}

\section {Related Work}

\textbf{Animal Identification:}
A variety of techniques have been proposed for animal identification \cite{si04BradfieldFrogPhoto, aje09ShorrocksNeckNet, fiz07SpeedSpotMatch}.
We outline the two most closely related in terms of feature
representation, matching, scoring, and results.

StripeSpotter \cite{icmr11LahiriStripeSpotter} bases recognition on features
called stripe-codes, two dimensional strings of binary values
designed to capture the typical stripe patterns of zebras.
Similarity between stripe-codes is measured by a
modified edit-distance dynamic-programming algorithm.  Queries are run
by measuring similarity to each database image individually and
returning the top matches.  On a database of 85 plains 
zebras, StripeSpotter achieves a median correct rank of 4.

Wild-ID \cite{11BoldgerWILDID} uses the original SIFT
\cite{ijcv04LoweSIFT} features and descriptors. It scores the query
image against each database image separately. For each feature in
the query image, the best match is found in each database image. These
matches are tested for inter-image consistency using randomly-sampled
triples of matches, a variation on more typical RANSAC methods.  The
score of each database image is the proportion of consistent triples.
On a database of 100 Wildebeest images, Wild-ID achieved a
false positive rate of $8.1 \times 10^{-4}$, with a false rejection
rate ranging from $.06-.08$ \cite{joae12MorrisonWetSeason}.  In our
one-vs-one algorithm we return to more common methods and introduce a
number of recent innovations to produce somewhat better performance.

Turning to the computer vision literature, \emph{instance recognition}
is the problem of finding the images in a database that match a query
image.  By contrast, \emph{category recognition} is the problem of
identifying the class an image belongs to --- e.g.\ car, person,
bicycle --- rather than identifying a particular instance of the
class. Our problem is closely related to instance recognition,
but we are more interested in identifying (labeling) the
object (animal) than in finding all images.

Research in instance recognition during the last decade has primarily
focused on large-scale search, starting with the development of visual
vocabularies \cite{iccv03SivicZissermanVideoGoogle} and fast matching
through vocabulary tree construction
\cite{cvpr06NisterSteweniusVocabTree}.  Typical
methods \cite{cvpr07PhilbinObjRetriev} build a visual vocabulary with
millions of words and represent images as visual-word vectors.
Scores between two images are computed using the TF-IDF
\cite{iccv03SivicZissermanVideoGoogle} weighted distance between
query and database visual-word vectors.  Searching large databases is
done efficiently by using an inverted file.  Spatial
verification re-ranks query results to produce a final
ranking of matches to the database image.  Recent innovations include
query expansion \cite{cvpr12ChumTotalRecallII}, Hamming
embedding, \cite{ijcv09JegouDouzeSchmidImprovBOF}, and feature augmentation
\cite{cvpr12ArandjelovicThreeThings}.
In our work, we adopt several of these features but ignore others
because (a) we are only interested in correctly identifying the animal (high precision) not
finding all matching images (total recall), and (b) we do not want to incur expensive preprocessing.

Category recognition methods have
tended to use dense sets of features and descriptors \cite{cvpr06SchmidBeyondBOF}, with
much smaller vocabularies than instance recognition methods.
By contrast, Boiman et al.\ \cite{cvpr08BoimanInDefenseNN} demonstrate the harmful
effect of quantization and describe an alternative non-quantized
approach. Given a set of query image descriptors, for every category, its match score is the sum of the square distances
between each query descriptor and the closet descriptor from that category, regardless of the
descriptor's parent image.
McCann and Lowe \cite{cvpr11McCannLNBNN} generalized this by simultaneously matching
each query descriptor to its $k$ nearest neighbors across all categories.  Our
one-vs-many algorithm applies this technique to animal
identification.

\section{One-Vs-One Matching}

We describe the one-vs-one algorithm first and then present the
modifications to create the faster one-vs-many algorithm.  Both
algorithms consist of five steps: 1) preprocessing, 2) matching, 3)
image scoring, 4) spatial reranking, and 5) label (animal name)
scoring.

\subsection{Preprocessing}

Preprocessing involves extracting features and building search data structures.
Before feature extraction, each database and query image is cropped to a rectangular region of interest (ROI) and
resized to a standard dimension, while preserving aspect ratio.

For image, $I$, either from the database or from the query, the
features are locations $\vect{x}$ of spatial and scale extrema of the
'Hessian-Hessian'~\cite{cvpr09PerdochEfficRep} operator applied to
$I$.  This operator is faster and more stable than the standard
'Hessian-Laplace'~\cite{ijcv05MikolajczykAffineRegion}, already
known to outperform the 'DoG' operator used in \cite{11BoldgerWILDID}.  An
elliptical shape is fit to the region
\cite{ijcv05MikolajczykAffineRegion}, represented as a matrix
$\matx{A}=[\begin{smallmatrix} a&0\\ b&c \end{smallmatrix}]$, with orientation angle, $\theta$~\cite{cvpr09PerdochEfficRep}.  Often $\theta$ is the aggregate of gradient
directions in the region surrounding $\vect{x}$, but instead, we assume
that the ``gravity vector'' is downward in each image and use this to
simply assign $\theta =0$. Removing the invariance to the gradient direction increases the
discrimination power of matching. For each region $(\vect{x},
\matx{A})$ we extract a RootSIFT~\cite{cvpr12ArandjelovicThreeThings} descriptor, $\vect{d} \in
\mathfrak{R}^{128}$, which is the
the component-wise square root of the usual SIFT descriptor vector
\cite{ijcv04LoweSIFT}.  Using the square root de-emphasizes the larger
components in a SIFT vector, making it more resistant to changes
between images.  The feature locations, elliptic regions and
descriptors for each image $I$ are stored in three
sets $\mathcal{X} = \{ \vect{x}_i \}, \mathcal{A} = \{ \matx{A}_i \}, \mathcal{D} = \{\vect{d}_i \}$.

The second part of preprocessing is the construction of a fast search data
structure.  For one-vs-one matching, we build a small forest of k-d
trees \cite{cvpr08HartleyKDTree} for the query image descriptors
using the VLFeat \cite{picm10VedaldiVLFEAT} library.  Since this is a one-vs-one
algorithm it is not necessary to build a search data structure for
each of the database images.

\subsection{One-vs-One Matching} \label{sec:matching1}

Let $I_D$ be the database image, with descriptor vectors
$\mathcal{D}_D$, and let $I_Q$ be the query image with descriptor
vectors $\mathcal{D}_Q$.  For each $\vect{d}_i \in \mathcal{D}_D$, the
two closest query image descriptors, $\{\vect{q}_{j},
\vect{q}_{j_2}\} \subset \mathcal{D}_Q$ are found (with high
probability) through a restricted, prioritized search of the k-d tree
forest.  Using the standard ratio-test \cite{ijcv04LoweSIFT}, a correspondence
is formed between the $i$-th feature of $I_D$ and the $j$-th feature
of $I_Q$ only if the ratio of nearest descriptor squared distances
 \begin{equation} \label{eqn:ratio}
 r_{i,j} = \frac{||\vect{d}_i - \vect{q}_{j_2}||^2}{||\vect{d}_i -  \vect{q}_{j}||^2}
 \end{equation}
exceeds some threshold, $t_{\tt ratio} = 1.6^2$.  All such
matches are collected into a set $\mathcal{M}_D = \{ (i, j, r_{i,j}) \}$.

\subsection{Initial Image Scoring} \label{sec:iis}

The initial image score between each database image $I_D$ and the
query image $I_Q$ depends simply on
the ratios stored in the match set, which encourages numerous distinctive correspondences.
\begin{equation} \label{eqn:sim}
 Sim(\mathcal{M}_D) = \sum_{(i, j, r_{i,j}) \in \mathcal{M}_D} r_{i,j}
\end{equation}


\subsection {Spatial Reranking} \label{sec:rerank1}

The initial image scores are computed without regard to
spatial constraints. To filter out any ``spatially inconsistent'' descriptors we implement the
standard RANSAC solution used in \cite{cvpr07PhilbinObjRetriev} and described in \cite{accv04ChumLORANSAC}.
We limit the computation
to the $K_{SR} = 50$ images with top initial image-scores, $Sim(\mathcal{M}_D)$.


To filter matches between $I_D$ and $I_Q$, we compute a set
of spatially consistent ``inliers'' for every match,
$(i, j, r) \in \mathcal{M}_D$ using the transformation between
 their affine shapes, $\matx{H} = \matx{A}_{i}^{-1}\matx{A}_{j}$.
More precisely, inliers are matches where the distance of a query feature, projected by $\matx{H}$, to its matching database feature
 is less than a spatial threshold, $t_{\tt sp}$. We set this to 10\% of $I_D$'s diagonal length.
The largest set of inliers is used to estimate a final homography. 
Inliers with respect to this homography become the final set of correspondences,
$\mathcal{M}'_D$.  Applying the image scoring function
$Sim(\mathcal{M}'_D)$ then determines the final, spatially-reranked
image score.


%

\subsection {Scoring of Labels} \label{sec:nmscore1}

Because we are interested in determining a query animal's label and
because there are typically multiple images associated with each label
in the database, we must convert the scores for each image into scores
for each label.  The simplest, obvious version, which we refer to as
\emph{image scoring}, is to select the label of the highest scoring re-ranked image.
 A second method, building off intuitions from
the Naive Bayes approach to category recognition
\cite{cvpr08BoimanInDefenseNN,cvpr11McCannLNBNN}, combines the scores from all
images with the same label.  This score, which we refer to as \emph{label scoring}, is computed in three steps: For each label, (a) aggregate match sets,
$\mathcal{M}'_D$, over all reranked images with this label, (b)
 for each query descriptor, remove all but the best match, and (c) use the value of $Sim$ applied to this new set as the score for this label.
%

\section {One-Vs-Many Matching}

Our one-vs-many algorithm is described relative to the one-vs-one algorithm.
These modifications give one-vs-many a logarithmic running time in the number
of database descriptors. We achieve this by deriving a new matching algorithm
for multi-image search from the Local Naive Bayes Nearest Neighbor (LNBNN) method \cite{cvpr11McCannLNBNN}.

\subsection {Preprocessing For Faster Search}

Feature extraction is identical to one-vs-one, but
computation of the efficient search data structure is very different.

Unlike one-vs-one, no search data structure is computed for the query image.
One-vs-many preprocessing computes \emph{one} nearest neighbor
data structure to index \emph{all} database descriptors.  This is
a small forest of k-d trees \cite{cvpr08HartleyKDTree}.   We refer to the aggregate
set of database descriptors as $\mathcal{D}_{\tt all}$.

\subsection{Descriptor Matching and Match Scoring} \label{sec:match2}

Similar to LNBNN \cite{cvpr11McCannLNBNN} for each query image descriptor $\vect{q} \in \mathcal{D}_Q$, the $k+1$
approximate nearest neighbors, $\{ \vect{d}_1, \vect{d}_2, ... \vect{d}_{k+1} \} \subset \mathcal{D}_{\tt all}$ are found by
searching the forest of k-d trees.  These are ordered by non-decreasing distance from $\vect{q}$.
Scores are computed for the first $k$ of these (and added to their
associated images) using the squared distance to the $(k+1)$-th neighbor as a normalizer.

Looking at this intuitively, if
there were just one database image, then we would be in the same
scenario as the one-vs-one matching, except with the roles of the
query and database images reversed, and our scoring mechanism should
reflect this.  On the other hand, in the usual
case where there are many more than $k$ database images, every query descriptor will only
generate scores for (at most) $k$ of them.  Matches for database
descriptors whose distances are beyond the $k$, just like matches
whose ratio scores are below the ratio cut-off in one-vs-one
matching, are considered non-distinct and (implicitly) given a score
of 0.

We present four scoring methods, using $\delta$ to denote the score
instead of $r$ (Equation~\ref{eqn:ratio}):

\begin{enumerate}
\item Applying the measure from LNBNN \cite{cvpr11McCannLNBNN} (modified to a maximization),  we define:
\begin{equation}
 \delta_{\tt LNBNN} (\vect{q},\vect{d}_{p},\vect{d}_{k+1}) = ||\vect{d}_{k+1} -  \vect{q}||^2  -  ||\vect{d}_{p} -  \vect{q}||^2
\end{equation}
This is just the difference in distances between the $p$-th and
$(k+1)$-th nearest neighbors. It tends towards 0 when the normalizing
feature is numerically close to the matching feature (hence not distinct).

\item We generalize the ratio score in Section~\ref{sec:matching1} to $k$ nearest neighbors:
\begin{equation}
  \delta_{\tt ratio} (\vect{q},\vect{d}_{p},\vect{d}_{k+1})  = \frac{||\vect{d}_{k+1} -  \vect{q}||^2}{||\vect{d}_{p} -  \vect{q}||^2}
\end{equation}
Note that when $k=1$, this is like the ratio test, but applied across
all database images, not just one.

\item We introduce a log-ratio score to drive the previous score to 0 when the matches are non-distinct:
\begin{equation}
  \delta_{\tt lnrat} (\vect{q},\vect{d}_{p},\vect{d}_{k+1})  = \ln(\frac{||\vect{d}_{k+1} -  \vect{q}||^2}{||\vect{d}_{p} -  \vect{q}||^2})
\end{equation}

\item As a sanity check we use match counting:
\begin{equation}
 \delta_{\tt count} (\vect{q},\vect{d}_{p},\vect{d}_{k+1}) = 1
\end{equation}
\end{enumerate}

If $\vect{q}$ is the $j$-th query image descriptor and if the $p$-th
closest descriptor is at descriptor index $i$ in its original database
image, $I_D$, then triple $(i,j,\delta(...))$ is appended to the match
set $\mathcal{M}_{D}$ for $I_D$.  Overall, if there are $M$ query
image descriptors, then $M k$ scores are generated and distributed
across all database images.

Once matching is complete, the initial image scoring
(Sec.~\ref{sec:iis}), spatial reranking (Sec.~\ref{sec:rerank1}) and
scoring of labels (Sec.~\ref{sec:nmscore1}) are all applied, just as in
the one-vs-one algorithm.

\subsection{Scaling to extremely large databases}

Our nearest neighbor methods are much simpler than the usual TF-IDF
scoring mechanism used in large-scale instance recognition, avoiding
the expensive off-line subcomputation of building the vocabulary and
inverted-file structure \cite{cvpr07PhilbinObjRetriev}.
If we can demonstrate that these are
effective, we will have taken a strong step toward a flexible, dynamic
database.  One drawback to our approach, however, is that each of the
original descriptor vectors is stored in memory as opposed to
quantized vectors.

To address this problem, we investigate replacing our k-d tree
indexing of database descriptor vectors with the product
quantization technique introduced by J\'{e}gou \etal~\cite{pami11JegouDouzeSchmidProdQuant}.
Product quantization is a method of computing
approximate nearest neighbors by splitting a (descriptor) vector
$\vect{z} = (z_1, z_2, \ldots, z_n)$ into $m$
subvectors $(\vect{u}_1, ..., \vect{u}_m)$, each of which is quantized
individually with a relatively small, separate vocabulary.  The
trick is that the representation is very small, but the implicit
vocabulary (number of distinct vectors) is enormous.  We use $m=16$
making each $\vect{u}_i$ length 8, and quantize at 128 words per
subvector.  This structure represents $5 \times 10^{33}$ distinct vectors and is 16 times smaller than the k-d
tree forest representation.  We explore the effects of this technique on match
quality.

\section{Experimental Results and Discussion}

All results were generated with MATLAB running on an i7-2600K 3.4GHz processor.
We used the open source C implementations of kd-trees in VLFeat \cite{picm10VedaldiVLFEAT} and feature extraction from Perdoch{} \cite{cvpr09PerdochEfficRep}.
We were provided with five datasets to test the HotSpotter algorithms.
We thank the Rubenstein team in Kenya for the large Grevy's and plains datasets as well as giraffes; Marcella Kelly for jaguars, and Juan David Gonz\'{a}lez Corredor for lionfish.
Summaries of each are provided in Table~\ref{tab:dbstats}, and examples of each are in Figure~\ref{fig:correctall}.

\subsection{Data Sets}

\newlength{\correctmatchheight}
\setlength{\correctmatchheight}{1.65in}

\begin{figure*}
\begin{center}
\begin{tabular}{c c c c c}
\includegraphics[height=\correctmatchheight]{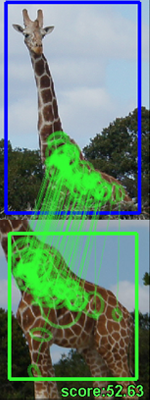}
&
\includegraphics[height=\correctmatchheight]{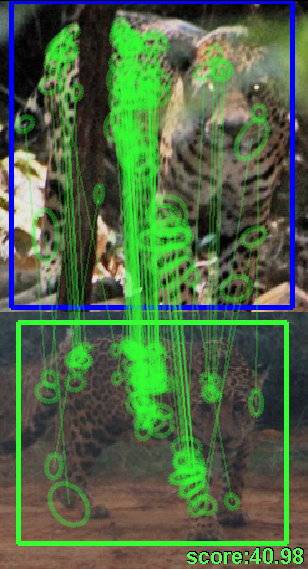}
&
\includegraphics[height=\correctmatchheight]{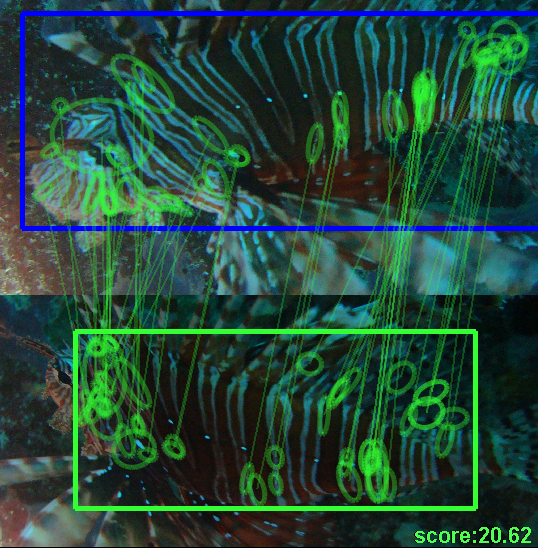}
&
\includegraphics[height=\correctmatchheight]{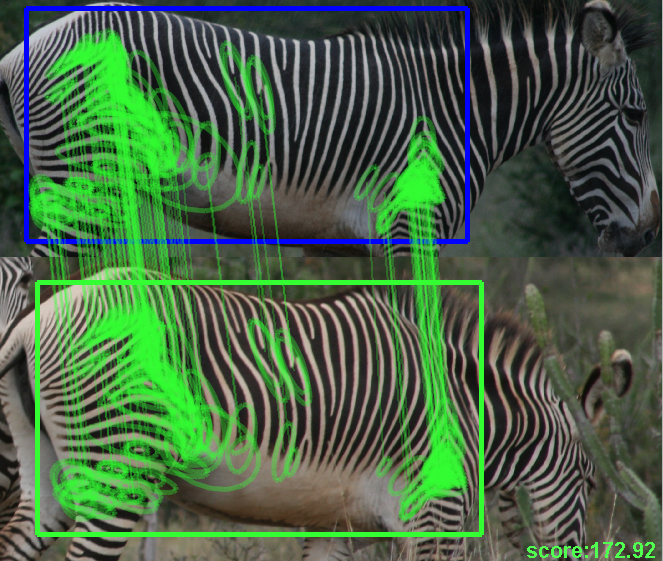}
&
\includegraphics[height=\correctmatchheight]{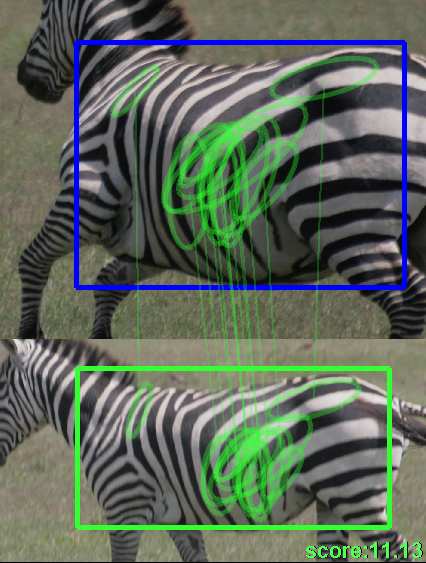}
\end{tabular}
\caption{\footnotesize{Correct matches. The top images are queries and the bottom images are top results. Wild-ID failed on jaguar and lionfish examples.}}
\label{fig:correctall}
\end{center}
\end{figure*}

\begin{table}
\centering
\footnotesize{
\begin{tabular}{|c | c c | c |}
\hline
\multirow{2}{*}{ Species }   & \multicolumn{2}{c|}{Number of}  &  Average Number of     \\
                             &  Images &   Labels              & Descriptors per Image  \\
\hline
Grevy's    & 1047 & 592 & $ 837.6$  \\
Plains     & 824  & 86  & $ 403.6$ \\
Jaguars    & 45   & 21  & $2660.5$ \\
Giraffes   & 45   & 15  & $1418.3$ \\
Lionfish  & 13   & 5   & $1183.3$ \\
\hline
\end{tabular}
\caption{\footnotesize{Dataset statistics}}
\label{tab:dbstats}
}
\end{table}

 \textbf{Grevy's zebras} 
are an endangered species with a small but stable population $\approx3,000$ \cite{williams2002status}.
They are characterized by a thin narrow stripe pattern, which produces many stable keypoints
along the shoulder and rump, and sometimes flank.
This dataset has been built from photos taken over several years and
contains images of animals as both juveniles and adults, making it our
most challenging set.

 \textbf{Plains zebras} 
 are abundant and widespread. Having
 coarse and thick stripe patterns, they would seem to generate a
 comparatively low number of distinguishing markings for recognition,
 and indeed they generate the fewest keypoints / descriptors .
Most images in this dataset were taken within seconds of each other, however,
leading to very little variation in either appearance or viewpoint.

 \textbf{Jaguars, giraffes, and lionfish} 
  are smaller datasets used to
  test HotSpotter's generalization to other species.  The jaguar images were
  taken from camera traps, are of low image quality, and sometimes
  show variations in viewpoint.  This difficulty is offset by a large
  number of distinguishing spots.  Giraffes and lionfish, forming smaller
  datasets here, are primarily taken from the same viewpoint.

\subsection{Experiments}

In every test configuration we take each database image with a known match and issued it as a query.
(We take care to ignore-self matches; thus removing the need to
reindex search structures.)  Knowledge of the correct matches allows us
to gather numerical results. (Interestingly, the software enabled us to catch and fix several ground-truthing errors.)

Results of running various configurations of HotSpotter on the larger
Grevy's and plains zebras sets are summarized in Tables ~\ref{tab:gz}
and ~\ref{tab:pz}. The first three columns of each table summarize the
algorithm configuration. Columns~4 and~5 give the number of queries
for which the correct label is not the first label returned for label
scoring (Column~4) and image scoring (Column~5) --- see
Sec.~\ref{sec:nmscore1}. This gives the overall accuracy of the
system.  Columns~6 and~7 give label scoring and image scoring for
queries where the returned rank is above 5.  This represents a
``usability'' measure since it is unlikely that a user will scan
beyond the first five labels.  Finally, the last column in the tables
is the query time per image in seconds.

Each row of Tables~\ref{tab:gz} and~\ref{tab:pz} gives a different
algorithm configuration, with the primary variations being between the
one-vs-one algorithm and the one-vs-many algorithm and, within the
one-vs-many algorithm, the number of neighbors returned, the scoring
function $\delta$, and whether or not product quantization is used.  In
addition, the default is for 50 images to be reranked, so the notation
$+$R0 and $+$RA in the first and last rows of the tables mean,
respectively, that no images and all images are reranked.  Finally, in all cases except those labeled $+$S,
Root-SIFT descriptors are used.

Results on the giraffes, leopard and lionfish datasets are provided in
Table~\ref{tab:others} in a much reduced form since these are easier
and less extensive.  The table includes a comparison to
Wild-ID~\cite{11BoldgerWILDID}.

\subsection{Results} \label{sec:results}

\begin{table}
\centering
\footnotesize{
\begin{tabular}{| c c c || c  c | c  c || c |}
\hline
 Algorithm:   & k & $\delta$ &\multicolumn{2}{c |}{Rank $>$ 1} & \multicolumn{2}{c ||}{Rank $>$ 5} & TPQ   \\
              &     &        & label & image                    & label & image                     & (sec) \\
\hline

1v1+R0 & 1 & $\tt ratio$ & 33 & 25 & 11 & 8 & 78.8\\
1v1 & 1 & $\tt ratio$ & 26 & 24 & 6 & 7 & 72.2\\
\hline
\hline
1vM & 1 & $\tt ratio$ & 33 & 36 & 8 & 5 & 4.0\\
1vM & 1 & $\tt lnrat$ & 32 & 34 & 6 & 5 & 4.0\\
1vM & 1 & $\tt count$ & 27 & 28 & 7 & 9 & 3.9\\
1vM & 1 & $\tt LNBNN$ & 40 & 47 & 8 & 9 & 4.0\\
1vM+PQ & 1 & $\tt LNBNN$ & 152 & 173 & 92 & 92 & 5.6\\
\hline
1vM & 5 & $\tt ratio$ & 37 & 34 & 5 & 6 & 3.4\\
1vM & 5 & $\tt lnrat$ & 34 & 33 & 5 & 5 & 3.8\\
1vM & 5 & $\tt count$ & 32 & 36 & 7 & 7 & 3.6\\
1vM & 5 & $\tt LNBNN$ & 39 & 41 & 4 & 5 & 3.7\\
1vM+PQ & 5 & $\tt LNBNN$ & 41 & 49 & 13 & 13 & 4.5\\
\hline
1vM & 10 & $\tt ratio$ & 38 & 35 & 6 & 6 & 3.5\\
1vM & 10 & $\tt lnrat$ & 33 & 33 & 4 & 5 & 3.8\\
1vM & 10 & $\tt count$ & 32 & 36 & 8 & 7 & 3.7\\
1vM & 10 & $\tt LNBNN$ & 39 & 41 & 4 & 5 & 4.0\\
1vM+PQ & 10 & $\tt LNBNN$ & 43 & 51 & 14 & 16 & 4.5\\
\hline
1vM & 20 & $\tt ratio$ & 38 & 36 & 6 & 6 & 3.4\\
1vM & 20 & $\tt lnrat$ & 32 & 31 & 5 & 5 & 3.7\\
1vM & 20 & $\tt count$ & 32 & 36 & 8 & 7 & 3.6\\
1vM & 20 & $\tt LNBNN$ & 38 & 41 & 6 & 6 & 3.8\\
1vM+PQ & 20 & $\tt LNBNN$ & 43 & 51 & 15 & 16 & 4.6\\
\hline
1vM+S & 1 & $\tt lnrat$ & 33 & 35 & 6 & 7 & 3.8\\
1vM+S & 1 & $\tt LNBNN$ & 40 & 45 & 7 & 7 & 3.7\\
\hline
1vM+R0 & 1 & $\tt lnrat$ & 34 & 36 & 7 & 5 & 1.5\\
1vM+RA & 1 & $\tt lnrat$ & 32 & 34 & 6 & 5 & 44.6\\
\hline
\end{tabular}
}
\caption{\footnotesize{Results on 657 Grevy's queries and different algorithm
  configurations.  The notation for the first column of the
  table indicates one-vs-one (1v1) matching, one-vs-many (1vM) matching,
  product quantization (PQ), whether no images were reranked
  (R0), whether all were reranked (RA), and if original SIFT
  descriptors (S) are used in place of Root-SIFT.  The column labeled $k$
  indicates the number of nearest neighbors scored for each query
  descriptor. The column labeled $\delta$ indicates the choice of
  match scoring function.  The results are show in the right five
  columns, summarizing the label rankings for label scoring and image
  scoring, and giving the query time.}}
\label{tab:gz}
\end{table}

\ifxetex
\else

\begin{table}
\centering
\footnotesize{
\begin{tabular}{| c c c || c  c | c  c || c |}
\hline
 Algorithm:   & k & $\delta$ &\multicolumn{2}{c |}{Rank $>$ 1} & \multicolumn{2}{c ||}{Rank $>$ 5} & TPQ   \\
              &     &        & label & image                    & label & image                     & (sec) \\
\hline

1v1+R0 & 1 & $\tt ratio$ & 9 & 5 & 5 & 3 & 41.0\\
1v1 & 1 & $\tt ratio$ & 7 & 4 & 3 & 3 & 38.2\\
\hline
\hline
1vM & 1 & $\tt ratio$ & 6 & 2 & 3 & 2 & 2.6\\
1vM & 1 & $\tt lnrat$ & 7 & 4 & 2 & 1 & 2.6\\
1vM & 1 & $\tt count$ & 6 & 3 & 2 & 2 & 2.6\\
1vM & 1 & $\tt LNBNN$ & 6 & 5 & 1 & 1 & 2.5\\
1vM+PQ & 1 & $\tt LNBNN$ & 25 & 29 & 15 & 15 & 2.7\\
\hline
1vM & 5 & $\tt ratio$ & 7 & 3 & 3 & 2 & 2.2\\
1vM & 5 & $\tt lnrat$ & 5 & 3 & 2 & 2 & 2.3\\
1vM & 5 & $\tt count$ & 8 & 4 & 3 & 2 & 2.4\\
1vM & 5 & $\tt LNBNN$ & 6 & 3 & 2 & 2 & 2.6\\
1vM+PQ & 5 & $\tt LNBNN$ & 7 & 7 & 3 & 4 & 3.1\\
\hline
1vM & 10 & $\tt ratio$ & 7 & 3 & 3 & 2 & 2.3\\
1vM & 10 & $\tt lnrat$ & 7 & 3 & 3 & 2 & 2.2\\
1vM & 10 & $\tt count$ & 8 & 4 & 3 & 2 & 2.5\\
1vM & 10 & $\tt LNBNN$ & 7 & 3 & 2 & 2 & 2.4\\
1vM+PQ & 10 & $\tt LNBNN$ & 7 & 7 & 2 & 3 & 3.3\\
\hline
1vM & 20 & $\tt ratio$ & 7 & 3 & 3 & 2 & 2.3\\
1vM & 20 & $\tt lnrat$ & 7 & 3 & 3 & 2 & 2.2\\
1vM & 20 & $\tt count$ & 8 & 4 & 3 & 2 & 2.5\\
1vM & 20 & $\tt LNBNN$ & 7 & 5 & 3 & 2 & 2.6\\
1vM+PQ & 20 & $\tt LNBNN$ & 7 & 6 & 2 & 3 & 3.3\\
\hline
1vM+S & 1 & $\tt lnrat$ & 7 & 4 & 3 & 2 & 2.4\\
1vM+S & 1 & $\tt LNBNN$ & 6 & 6 & 2 & 2 & 2.5\\
\hline
1vM+R0 & 1 & $\tt lnrat$ & 7 & 5 & 4 & 3 & 0.7\\
1vM+RA & 1 & $\tt lnrat$ & 6 & 4 & 3 & 2 & 30.4\\
\hline
\end{tabular}
}
\caption{\footnotesize{Results on 819 plains zebras queries using the same notation
  at in Table~\ref{tab:gz}.}}
\label{tab:pz}
\end{table}

\begin{table}
\centering
\footnotesize{
\begin{tabular}{| l c c || c  c | c  c || c |}
\hline
 Algorithm:   & k & $\delta$ &\multicolumn{2}{c |}{Rank $>$ 1} & \multicolumn{2}{c ||}{Rank $>$ 5} & TPQ   \\
              &     &        & label & image                    & label & image                     & (sec) \\
\hline
\hline
 \multicolumn{8}{| c |}{Dataset: Giraffes} \\
 \hline
 1v1+RA      & 1   & $\tt ratio$ & 1 & 0 & 1 & 0 & 3.3 \\
 1vM+RA      & 1   & $\tt ratio$ & 1 & 0 & 0 & 0 & 1.1 \\
 Wild-ID     & --  &    --     &-- & 4 &-- & 1 & 0.5 \\
\hline
\hline
 \multicolumn{8}{| c |}{Dataset: Jaguars} \\
 \hline
 1v1+RA      & 1   & $\tt ratio$ & 1  & 0 & 0  & 0 & 6.2 \\
 1vM+RA      & 1   & $\tt ratio$ & 1  & 0 & 0  & 0 & 2.0 \\
 Wild-ID     & --  &    --     & -- & 3 & -- & 2 & 2.6 \\
\hline
\hline
 \multicolumn{8}{| c |}{Dataset: Lionfish} \\
 \hline
 1v1+RA      & 1   & $\tt ratio$ & 0  & 0 & 0 & 0 & 0.7 \\
 1vM+RA      & 1   & $\tt ratio$ & 0  & 0 & 0 & 0 & 0.7 \\
 Wild-ID     & --  &    --     & -- & 1 &-- & 0 & 1.5 \\
 \hline
\end{tabular}
}
\caption{\footnotesize{Results on giraffes, jaguars, and lionfish with 38, 35, and
  13 queries, respectively.}}
\label{tab:others}
\end{table}
\fi
Overall, the results are quite strong. We achieve perfect results for
the giraffe, jaguar, and lionfish datasets.  On the Grevy's
data set, in the best configuration, HotSpotter produces correct
top-ranked labels for 95\% of the queries, and a top-five label for
98\%.  For the plains zebras dataset, these numbers are both well over
99\%.

More specific observations about the results are as follows:

\begin{itemize}
\item One-vs-many is on par with one-vs-one matching, but many times
  faster.
\item Product quantization rankings are substantially improved by
  increasing $k$ from 1 to 5.  We attribute this improvement to
   an increased $k$ overcoming some of the
  effects of quantization; beyond $k=5$ there is no improvement.  For
  the other algorithms and configurations, setting $k=1$, which is
  effectively a multi-image generalization of Lowe's ratio test, works
  as well as any value of $k$.
\item The choice of scoring function $\delta$ does not make a
  substantial difference in the algorithm, and even the simple
  counting function works well.
\item Numerically, the difference between label scoring and image scoring
  (Sec.~\ref{sec:nmscore1} is minor, with small variations in either
  direction occurring with different algorithm configurations.
  Interestingly, it appears that one place label scoring does make a
  difference is matching images of foals as they grow and change
  toward adulthood.
\item Spatial reranking only marginally improves rankings, more so for
  the one-vs-one algorithm.
\item Unlike results reported in \cite{cvpr12ArandjelovicThreeThings},
  Root-SIFT has no significant advantage over standard SIFT for our
  data sets.
\end{itemize}
Combined, these observations suggest that the biggest reason for
success is the shear number of matches from distinctive regions of an animal.
The descriptors (``hotspots'') tend to be
distinctive across many different animals, effectively creating a
signature for that animal. This appears to be true for all five
species that we experimented with.  Even for the simple counting
algorithm, the insensitivity to $k$ is because the incorrect descriptor matches are
uncorrelated across in the database. This slightly increases the
noise in the scoring, but not enough to challenge the correct scores.

As shown in Table~\ref{tab:others}, Wild-ID is faster than the
one-vs-one version of HotSpotter, but missed matches that HotSpotter
finds. Two cases where Wild-ID fails are seen in Figure
\ref{fig:correctall}.  Since Wild-ID is structurally similar
to  HotSpotter in its one-vs-one configuration, we attribute these
successes primarily to the denser, affine-invariant keypoints
and descriptors. Although Wild-ID is faster than one-vs-one, this is largely due
to it being multi-threaded and implemented in Java instead of MATLAB.

\subsection{Failure Cases}

\begin{figure}
\begin{center}
\begin{tabular}{c}
\includegraphics[width=.9\linewidth]{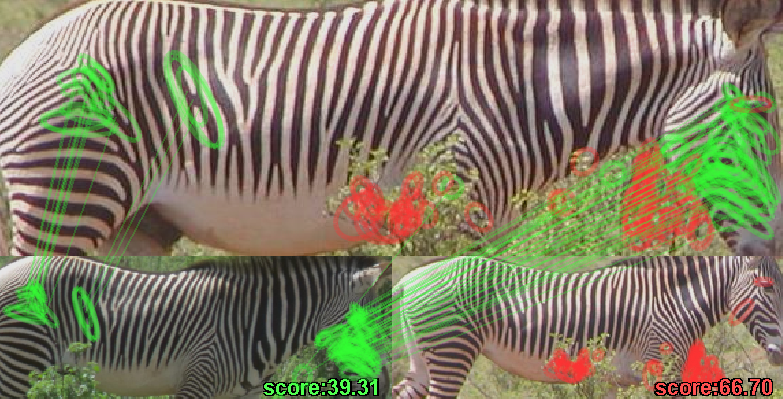}
\\
\includegraphics[width=.9\linewidth]{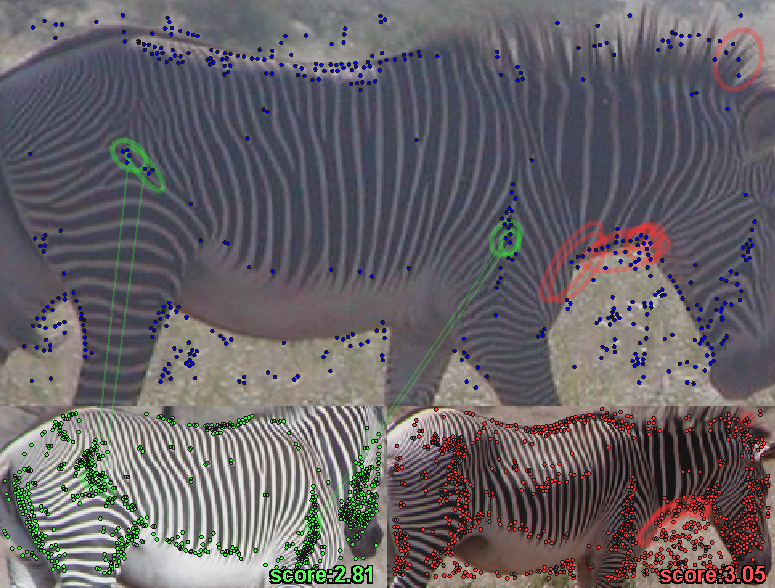}
\\
\includegraphics[width=.9\linewidth]{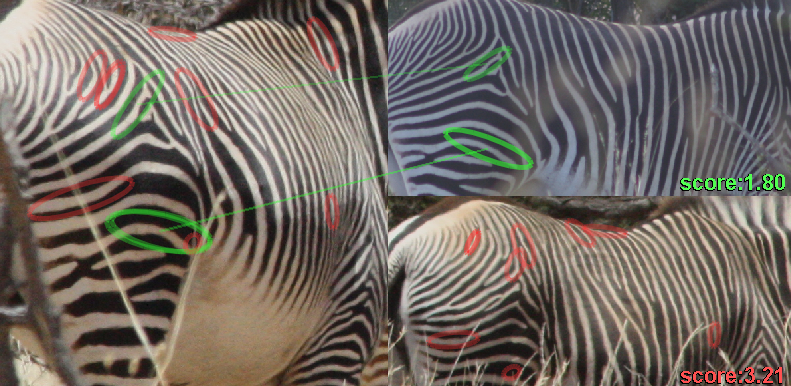}
\end{tabular}
\caption{\footnotesize{Three example failures.  In the top example, the same
  background is seen (red matches) in two images showing different zebras.  In
  the middle, the viewpoints are too different between the query image
  and the correct database image (green matches), and a different
  image is the top match.  On the bottom, the poor illumination of
  the query image produces too few keypoints (small circles in the
  images) for correct matching.}}
\label{fig:problem}
\end{center}
\end{figure}

Further insight into HotSpotter can be gained through an analysis of
failures.  We closely analyzed the 34 Grevy's queries which produce a rank greater than
one for one-vs-many with $k=5$ and $\delta={\textrm{\tt lnrat}}$.
In 2 cases the database contained a mislabeling, and in 5 cases the
ROI covered two animals and matched the unlabeled animal. In 13 other cases,
the background scenery was matched instead of the
foreground zebra region.  A classifier that weights a descriptor
according to how likely it is to be a zebra should fix this problem.
Of the remaining 14 cases, 8 failed because of substantial pose variations between query
and database zebras, and the rest because of image quality, including focus,
resolution and contrast.  Examples are shown in Figure \ref{fig:problem}.

\section{Conclusion}

We have presented HotSpotter, an algorithm for fast,
reliable, multi-species animal identification based on
extracting and matching keypoints and descriptors.  In addition to the
construction and testing of the overall system, the primary technical
contribution of our work is the development of a fast and scalable
one-vs-many scoring mechanism which outperforms current brute force
one-vs-one image comparison methods in both speed and accuracy.  The
key to the success of HotSpotter is the use of viewpoint invariant
descriptors and a scoring mechanism that emphasizes the most
distinctiveness keypoints and descriptors by allowing only the $k$
nearest neighbors of any descriptor to participate in the scoring.
This has been borne out on experiments with Grevy's zebras, plains
zebras, giraffes, leopards and lionfish.

From the perspective of the application, the failures are generally
manageable through human interaction to eliminate problems due to (a)
ROIs that overlap multiple animals, (b) matching
against background, and (c) poor quality images. Additional algorithms
could help identify and eliminate these problems as well.  Other
failures --- those due to viewpoint variations --- are the most difficult to handle. The
best way to address this is to incorporate multiple images and
viewpoints into the database for each animal.

Looking toward future experimental work, we will  
  apply the software to a wider variety of species, which will test the limit 
  recognition based on distinctive feature matching. From the perspective 
of both the algorithm and the practical application,
 the next major step for HotSpotter is to build and manage a dynamically
constructed database of animal images and labels.  This is the focus of
our ongoing effort.




{\small
\bibliographystyle{wacv2013-author-kit/latex/ieee}
\bibliography{main}

}

\end{document}